# Handwritten Farsi Character Recognition using Artificial Neural Network

Reza gharoie ahangar, Azad University.
The master of business administration
of Islamic Azad University - Babol branch &
Membership of young researcher club, Iran.
r.gharoie@gmail.com
.

Mohammad Farajpoor Ahangar, Babol University.
University of medical sciences of Babol, Iran. &
Membership of young researcher club, Iran
fraj.ahangar@yahoo.com

*Abstract*-Neural Networks are being used for character recognition from last many years but most of the work was confined to English character recognition. Till date, a very little work has been reported for Handwritten Farsi Character recognition. In this paper, we have made an attempt to recognize handwritten Farsi characters by using a multilayer perceptron with one hidden layer. The error backpropagation algorithm has been used to train the MLP network. In addition, an analysis has been carried out to determine the number of hidden nodes to achieve high performance of backpropagation network in the recognition of handwritten Farsi characters. The system has been trained using several different forms of handwriting provided by both male and female participants of different age groups. Finally, this rigorous training results an automatic HCR system using MLP network. In this work, the experiments were carried out on two hundred fifty samples of five writers. The results showed that the MLP networks trained by the error backpropagation algorithm are superior in recognition accuracy and memory usage. The result indicates that the backpropagation network provides good recognition accuracy of more than 80% of handwritten Farsi characters.

## I. INTRODUCTION

Handwritten character recognition is a difficult problem due to the great variations of writing styles, different size and orientation angle of the characters. Maybe among different branches of handwritten character recognition, it is easier to recognize Persian alphabets and numerals than Farsi characters. There have been only a few attempts made in the past to address the recognition of handwritten Farsi Characters [2].Character recognition is an area of pattern recognition that has been the subject of considerable research during the last some decades. Many reports of character recognition of several languages, such as Chinese [7], Japanese, English [3, 14, 15], Arabic [10, 11] and Farsi [5] have been published but still recognition of handwritten Farsi characters using neural networks is an open problem. Farsi is a first official Iranian language and it is widely used in many Iranian states. In many Iranian offices such as passport, bank, sales tax, railway, embassy, etc.: the Farsi languages are used. Therefore, it is a great importance to develop an automatic character recognition system for Farsi language [5].In this paper, we exploit the use of neural networks for off-line Farsi handwriting recognition. Neural networks have been widely used in the field of handwriting recognition [6, 8]. The present work describes a system for offline recognition of Farsi script, a language widely spoken in Iran. In this paper, we present MLP network for the handwritten Farsi character recognition and develop an automatic character recognition system using this network.

## II. FARSI LANGUAG

Farsi, which is a Iranian language, is one of the oldest languages in the world. There are 32 characters in Farsi language and is written from right to left. A set of handwritten Farsi character is shown in Figure1.

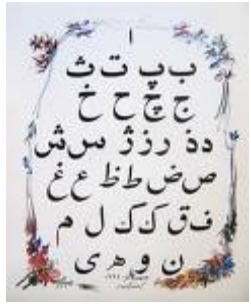

Figure1. A set of Handwritten Farsi Characters [5]

## III. PREPROCESSING

The handwritten character data samples were acquired from various students and faculty members both male and female of different age groups. Their handwriting was sampled on A4 size paper. They were scanned using flat-bed scanner at a resolution of 100dpi and stored as 8-bit grey scale images. Some of the common operations performed prior to recognition are smoothing, thresholding and skeletonization [2].

A. Image Smoothing

The task of smoothing is to remove unnecessary noise present in the image. Spatial filters could be used. To reduce the effect of noise, the image is smoothed using a Gaussian filter [2].

B. Skeletonization

We have initialized the mouse in graphics mode so that a character can be directly written on screen. The skeletonization process has been used to binary pixel image and the extra pixels which do not belong to the backbone of the character, were deleted and the broad strokes were reduced to thin lines. Skeletonization process is illustrated in Figure2. A character before and after skeletonization is shown in Figure 2a and 2b respectively [1].

C. Normalization

After skeletonization process, we used a normalization process, which normalized the character into 30x30-pixel character and it was shifted to the left and upper corner of pixel window. The final skeltonized and normalized character is shown in Figure 2c, which was used as an input of the neural network. The Skeletonization and Normalization process were used for each character [1].

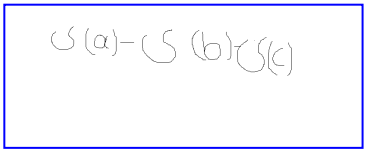

Figure2. Skeletonization and Normalization process of a Farsi characters [1].

IV. NEURAL NETWORK

A. Recognition

Recognition of handwritten letters is a very complex problem. The letters could be written in different size, orientation, thickness, format and dimension. These will give infinity variations. The capability of neural network to generalize and be insensitive to the missing data would be very beneficial in recognizing handwritten letters. The proposed Farsi handwritten character recognition system uses a neural network based approach to recognize the characters. Feed forward Multi Layered Perceptron (MLP) network with one hidden layer trained using back-propagation algorithm has been used to recognize handwritten Farsi characters [1, 2].

B. Structure Analysis of Backpropagation Network

The recognition performance of the Backpropagation network will highly depend on the structure of the network and training algorithm. In the proposed system, Backpropagation algorithm has been selected to train the network. It has been shown that the algorithm has much better learning rate. The number of nodes in input, hidden and output layers will determine the network structure. The best network structure is normally problem dependent, hence structure analysis has to be carried out to identify the optimum structure [2]. We have used multilayer perceptron trained by Error Backpropagation (EBP) neural network classification technique [1]. A brief description of this network is presented in this section.

C. Multilayer Perceptron Network

The Multilayer Perceptron Network may be formed by simply cascading a group of single layer perceptron network; the output of one layer provides the input to the subsequent layer [16, 17]. The MLPN with the EBP algorithm has been applied to the wide variety of problems [1-17]. We have used a two-layer perceptron i.e. single hidden layer and output layer. A structure of MLP network for Farsi character recognition is shown in Figure3.

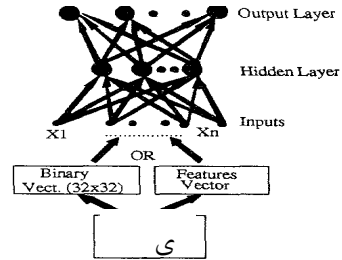

Figure3. Multilayer Perceptron Network [1]

The activation function of a neuron j can be expressed as:

$$F_j(x) = 1/(1+e^{-net}), \text{ where } net = \sum W_{ij}O_i \quad (1)$$

Where $O_i$ is the output of unit i, $W_{ij}$ is the weight from unit i to unit j. The generalized delta rule algorithm [1, 16, and 17] has been used to update the weights of the neural network in order to minimize the cost function:

$$E = \frac{1}{2}(\sum(D_{pk}-O_{pk}))^2 \quad (2)$$

Where $D_{pk}$ and $O_{pk}$ are the desired and actual values, respectively, of the output unit k and training pair p. Convergence is achieved by updating the weights by using the following formulas:

$$W_{ij}(n+1) = W_{ij}(n) + \Delta W_{ij}(n) \quad (3)$$
$$\Delta W_{ij}(n) = \eta \delta X_J + \alpha(W_{ij}(N)-W_{ij}(n-1)) \quad (4)$$

Where $\eta$ is the learning rate, $\alpha$ is the momentum, $W_{ij}(n)$ is the weight from hidden node i or from an input to node j at nth iteration, $X_i$ is either the output of unit i or is an input, and $\delta_j$ is an error term for unit j. If unit j is an output unit, then

$$\delta_j = O_j(1-O_j)(D_j-O_j) \quad (5)$$

If unit j is an internal hidden unit, then

$$\delta_j = O_j(1-O_j)\sum \delta_k W_{kj} \quad (6)$$

V. EXPERIMENTAL RESULT

A. Character Database

We have collected 250 samples of handwritten Farsi characters written by ten different persons 25 each directly on screen. We have used 125 samples as a training data (training set) and remaining 125 samples as a test data (test set).

B. Character Recognition with MLPN

We have implemented an automatic handwritten Farsi character recognition system using Multi- Layer Perceptron

(MLP) network in C/C++ language. A complete system may be shown in Figure 4.

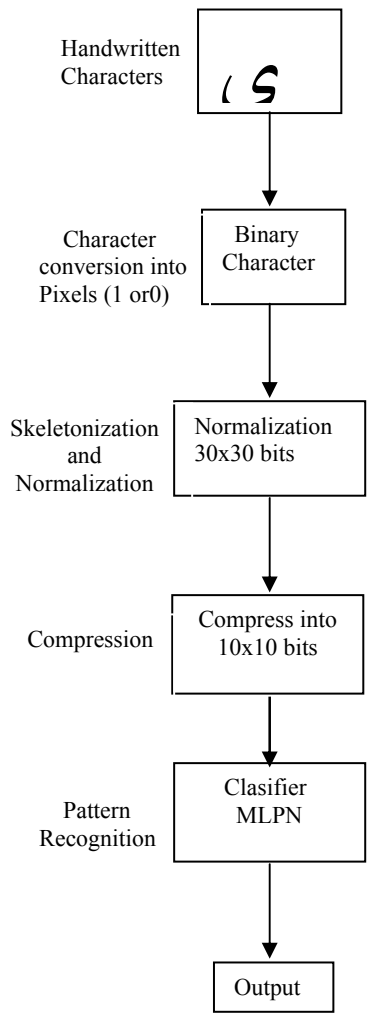

Figure4: A System for Farsi Character Recognition

We have initialized the mouse in graphics mode due to which we can write directly on screen with mouse. Once character has been written on screen, it is converted into binary pixels. After that, we perform a normalization process that converts the character represented in binary form into 30x30 bits. In next step, we compress the 30x30 bits into 10x10 bits. After that we apply neural network classifier in order to recognize the Farsi character. We have coded the Farsi character and made the Backpropagation neural network to achieve the coded value i.e. Supervised learning. For example for the character (ي), we have code 1 and made the network to achieve this value by modifying the weight functions repeatedly. Each MLP network uses two-layer feedfomard network [4] with nonlinear sigmoidal functions. Many experiments with the various numbers of hidden units for each network were carried out. In this paper, we have taken one hidden layer with flexible number of neurons and output layer with 05 neurons because we have collected the samples from (ي) to(الف). The network has been trained using the EBP algorithm as described in Section 4 and was trained until mean square error between the network output and desired output falls bellow 0.05. The weights were updated after each pattern presentation. The learning rate and momentum were 0.2 and 0.1 respectively. The results are shown in following Table1.

| Input of the MLPN | No. of hidden units | No. of iteration | Training time (s) | Recognition Accuracy (%) | |
|---|---|---|---|---|---|
| | | | | Training Data | Test Data |
| 30x30 | 12 | 200 | 1625 | 100 | 80 |
| | **24** | **200** | **3125** | **100** | **85** |
| | 36 | 200 | 4750 | 100 | 80 |

Table 1.Results of handwritten Farsi characters using MLPN

This table indicates network results for different states. For MLP network with 20,24 and 36 neurons in middle layer and with equal iteration, you will observe different quantities for predicting precision, and we see that network with 24 neurons give us response equal with 85 in test series, which is the most desirable answer than the others.

## VI. DISCUSSION

The results presented in previous subsections shows that 24 hidden units give the best performance on training set and test set for MLP network. The MLP networks takes longer training time because they use iterative training algorithm such as EBP, but shorter classification time because of simple dot-product calculations.

Here we should point to this issue that network with more neurons in the middle layer is not a better measure for network functioning, as we see that with increased neurons of middle layer, there is no improvement in the response of network.

## VII. CONCLUSION

In this paper, we have presented a system for recognizing handwritten Farsi characters. An experimental result shows that backpropagation network yields good recognition accuracy of 85%.

The methods described here for Farsi handwritten character recognition can be extended for other Iranian scripts by including few other preprocessing activities. We have demonstrated the application of MLP network to the handwritten Farsi character recognition problem. The skeletonized and normalized binary pixels of Farsi cliaracters were used as the inputs of the MLP network.

In our further research work, we would like to improve the recognition accuracy of network for Farsi character recognition by using more training samples written by one person and by using a good feature extraction system. The training time may be reduced by using a good feature extraction technique and instead of using global input, we may use the feature input along with other neural network classifier.